\documentclass[letterpaper, 10pt, conference]{style/ieeeconf}    %

\makeatletter
\let\NAT@parse\undefined
\makeatother

\usepackage{dblfloatfix}

\usepackage[numbers,sectionbib,sort&compress]{natbib}
\usepackage{bm}
\usepackage{gensymb}
\usepackage{graphicx}
\usepackage{amsmath}
\usepackage{amssymb}
\usepackage{algorithm}
\usepackage[noend]{algpseudocode}
\usepackage{subcaption}
\usepackage{amsfonts}
\usepackage{siunitx}
\usepackage{booktabs}
\usepackage{makecell}
\usepackage{multirow}
\usepackage{upgreek}
\usepackage[font=small]{caption}
\usepackage[export]{adjustbox}
\usepackage{tikz}
\usepackage{sidecap} \sidecaptionvpos{figure}{c}
\usepackage{xspace}
\usepackage{hyperref}
\usepackage{threeparttable}
\usepackage{tablefootnote}
\usepackage{arydshln}

\hypersetup{colorlinks,breaklinks,
linkcolor=[rgb]{0.5,0.,0.},
citecolor=[rgb]{0.000,0.427,0.173},
urlcolor=[rgb]{0.031,0.318,0.612}}

\usepackage[nameinlink]{cleveref}
\usepackage{color}
\definecolor{CommentPink}{rgb}{1,0.2,0.5}
\definecolor{CommentBlue}{rgb}{0,0,1}
\definecolor{CommentGreen}{rgb}{0,1,0}

\usepackage[printonlyused,withpage,nolist,nohyperlinks]{acronym}

\def\secref#1{Sec.~\ref{#1}}
\def\figref#1{Fig.~\ref{#1}}
\def\tabref#1{Tab.~\ref{#1}}
\def\eqref#1{Eq.~(\ref{#1})}

\begin{acronym}

\acro{2D}{two-dimensional}

\acro{3D}{three-dimensional}

\acro{AHRS}{attitude and heading reference system}

\acro{AUV}{autonomous underwater vehicle}

\acro{CPP}{Chinese Postman Problem}

\acro{DoF}{degree of freedom}
\acrodefplural{DoF}[DoFs]{degrees of freedom}

\acro{DVL}{Doppler velocity log}

\acro{FSM}{finite state machine}

\acro{IMU}{inertial measurement unit}

\acro{LBL}{Long Baseline}

\acro{MCM}{mine countermeasures}

\acro{MDP}{Markov decision process}
\acrodefplural{MDP}[MDPs]{Markov decision processes}

\acro{POMDP}{Partially Observable Markov Decision Process}
\acrodefplural{POMDP}[POMDPs]{Partially Observable Markov Decision Processes}

\acro{PRM}{Probabilistic Roadmap}
\acrodefplural{PRM}[PRM]{Probabilistic Roadmaps}

\acro{ROI}{region of interest}
\acrodefplural{ROI}[ROIs]{regions of interest}

\acro{ROS}{Robot Operating System}

\acro{ROV}{remotely operated vehicle}

\acro{RRT}{Rapidly-exploring Random Tree}
\acrodefplural{RRT}[RRTs]{Rapidly-exploring Random Trees}

\acro{SLAM}{Simultaneous Localisation and Mapping}

\acro{SSE}{sum of squared errors}

\acro{STOMP}{Stochastic Trajectory Optimization for Motion Planning}

\acro{TRN}{Terrain-Relative Navigation}

\acro{UAV}{unmanned aerial vehicle}

\acro{USBL}{Ultra-Short Baseline}

\acro{IPP}{informative path planning}

\acro{FoV}{field of view}
\acrodefplural{FoV}[FoVs]{fields of view}

\acro{CDF}{cumulative distribution function}

\acro{ML}{maximum likelihood}

\acro{RMSE}{Root Mean Squared Error}
\acro{MLL}{Mean Log Loss}

\acro{GP}{Gaussian Process}
\acrodefplural{GP}[GPs]{Gaussian Processes}

\acro{KF}{Kalman Filter}

\acro{IP}{Interior Point}
\acro{BO}{Bayesian Optimization}

\acro{SE}{squared exponential}

\acro{UI}{uncertain input}

\acro{MCL}{Monte Carlo Localisation}
\acro{AMCL}{Adaptive Monte Carlo Localisation}

\acro{SSIM}{Structural Similarity Index}

\acro{MAE}{Mean Absolute Error}
\acro{RMSE}{Root Mean Squared Error}
\acro{AUSE}{Area Under the Sparsification Error curve}

\end{acronym}

\newcommand{\etal}{\textit{et~al}.}

\IEEEoverridecommandlockouts                           %
\usepackage{array}

\title{3D Lidar Reconstruction with Probabilistic Depth Completion\\ 
for Robotic Navigation}
\author{Yifu Tao\textsuperscript{1}, Marija Popovi\'{c}\textsuperscript{2}, Yiduo Wang\textsuperscript{1}, Sundara Tejaswi Digumarti\textsuperscript{1}, Nived Chebrolu\textsuperscript{1}, and Maurice Fallon\textsuperscript{1}
    \thanks{This research is supported by the EPSRC ORCA Robotics Hub 
    	(EP/R026173/1), the EU H2020 Project MEMMO (780684), and partially 
    	supported by the Deutsche Forschungsgemeinschaft (DFG, German Research 
    	Foundation) under Germany’s Excellence Strategy - EXC 2070 – 390732324. 
    	M. Fallon is supported by a Royal Society University Research 
    	Fellowship. For the purpose of Open Access, the author has applied a CC 
    	BY public copyright licence to any Author Accepted Manuscript (AAM) 
    	version arising from this submission.}
\thanks{\textsuperscript{1} These authors are with the Oxford Robotics Institute, University of Oxford, UK.
	{\tt\small \{yifu, ywang, tejaswid,nived, mfallon\}@robots.ox.ac.uk}
        \textsuperscript{2} This author is with the Cluster of Excellence PhenoRob, Institute of Geodesy and Geoinformation, University of Bonn.
    \tt\small mpopovic@uni-bonn.de
}
}

\begin{document}

\maketitle
\thispagestyle{empty}
\pagestyle{empty}

\begin{abstract}
Safe motion planning in robotics requires planning into space which
has been verified to be free of obstacles.
However, obtaining such environment representations using lidars
is challenging by virtue of the sparsity of their depth measurements.
We present a learning-aided 3D lidar reconstruction framework 
that upsamples sparse lidar depth measurements with the aid of 
overlapping camera images so as to generate denser reconstructions
with more definitively free space than can be achieved with the 
raw lidar measurements alone.
We use a neural network with an encoder-decoder structure to predict 
dense depth images along with depth uncertainty estimates which are fused using a volumetric mapping system.
We conduct experiments on real-world outdoor datasets captured using
a handheld sensing device and a legged robot. 
Using input data from a 16-beam lidar mapping a building network, 
our experiments showed that the amount of estimated free space was 
increased by more than 40\% with our approach.
We also show that our approach trained on a synthetic dataset generalises
well to real-world outdoor scenes without additional fine-tuning.
Finally, we demonstrate how motion planning 
tasks can benefit from
these denser reconstructions. 
\end{abstract}

\section{Introduction} \label{S:introduction}

Dense 3D reconstruction is a key task in a range of robotic
applications including exploration, industrial inspection and
obstacle avoidance. Lidar is the dominant sensor used for mapping outdoor
environments as it provides accurate long-range depth measurements.
However, lidar readings are sparse, especially when compared to depth images obtained from RGB-D 
cameras.
For instance, the reprojection of a Velodyne 64-beam lidar only covers
around 6\% of the pixels of an overlapping camera image in the KITTI 
dataset~\cite{sparseconv}. This drops to about 1.6\% for a 16-beam 
sensor. 

Mapping with sparse lidar sensors results in incomplete reconstructions which in turn causes
path planning algorithms to either fail or be inefficient due to the limited free space detected.
This becomes a critical issue, particularly in robotic applications where path planning is performed online and
a reconstruction which minimises unobserved space is necessary for safe operation.  While  high-end lidars (e.g. 64 or 128 beams)
provide denser point clouds of the environment, their prices are typically restrictive for low-cost mobile robot platforms
such as delivery robots. 

\begin{figure}[t!]
	\centering
	\includegraphics[width=0.99\columnwidth]{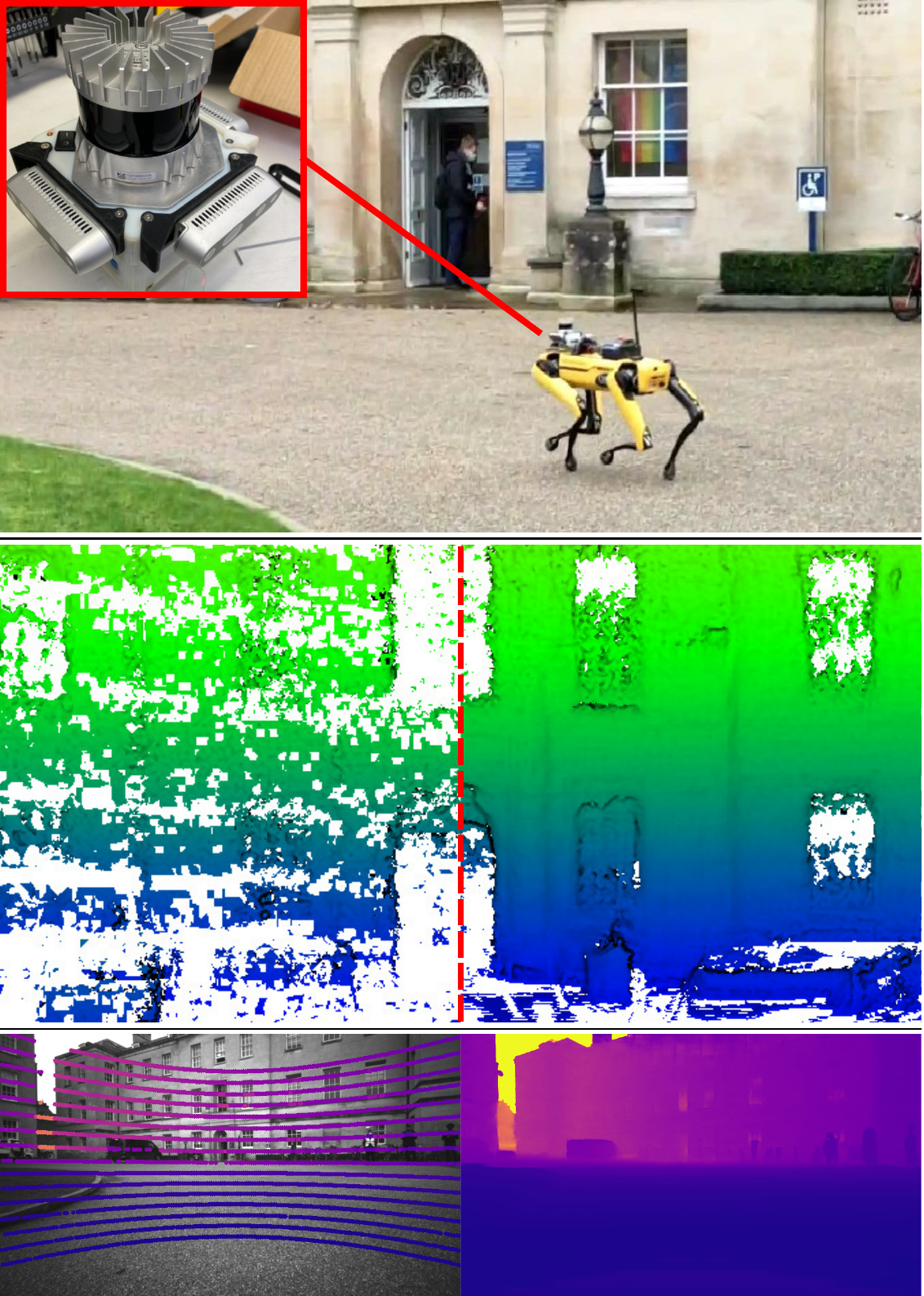}
	\caption{
		Depth completion and mapping results from our Maths Institute dataset.
		Top: A view of a building scanned by the legged robot. Inset shows the scanning device with a lidar and 3 cameras. 
		Middle: 3D reconstruction of the same building using the sparse 16-beam lidar (left) \& completed depth images estimated by our approach (right).
		Bottom: Forward camera image with sparse lidar overlay, our input (left) and a completed depth image output by our approach (right). 
		Video: \url{https://tinyurl.com/depth-completion}}
\label{fig:intro}
\vspace{-4mm}
\end{figure}

\begin{figure*}[!t]
    \centering
    \includegraphics[width=1.9\columnwidth]{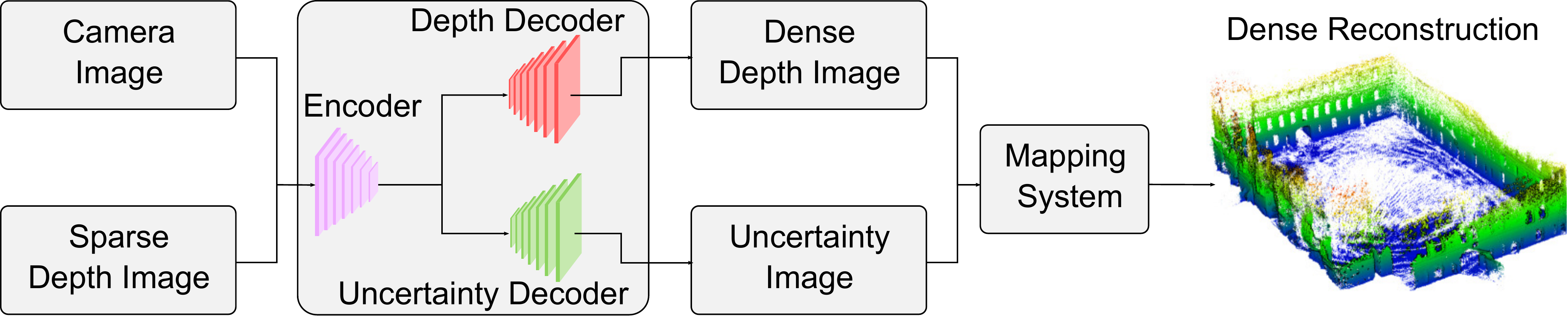}
    \caption{System pipeline: A camera image and a sparse lidar depth image are the inputs to the network. The network predicts
        dense depth images along with corresponding uncertainty values.
        These outputs are fed into a volumetric mapping system to
        obtain a 3D reconstruction. Our experiments used networks whose architectures are based on~\cite{sparse2dense, Ma2019, wofk2019fastdepth}.}
    \label{fig:pipeline}
    \vspace{-3mm}
\end{figure*}

To address the sparsity of lidar sensors, many works use texture information from an overlapping (monocular) camera 
to infer a complete depth image for every pixel of the camera, which 
is known as \textit{depth completion}. Recent learning-based approaches~\cite{cspn++,penet} have
made great progress in improving accuracy.
However, the completed depth is still subject to outliers, which would then generate distorted reconstruction and false positive free space volumes in regions that are truly unknown or occupied.

In this work, we propose a pipeline for large-scale mapping with 
probabilistic lidar depth completion which uses sparse 
lidar sensing and one or more overlapping camera images. Specifically, the neural 
network takes a pair
of camera and sparse lidar depth images as input, and predicts a 
dense depth image along with an associated depth uncertainty image.
These two predictions are then fused in our volumetric mapping system.
This approach can reconstruct a more complete structure
and detect more correct free space thanks to this high
density completed depth image. The accuracy of both the
surface reconstruction and the free space
is improved by rejecting predicted depth that has high
uncertainty predicted by the network. 
We deploy the proposed system on a legged robot platform. Unlike most lidar depth completion
works which typically use a single monocular camera image having only a small overlap with the lidar,
we use a three-camera setup (left, forward and right facing) to give about 270~\degree overlap with the lidar.

In summary, the contributions of this work are:
\begin{enumerate}
	\item Extension of a depth completion framework to incorporate uncertainty predictions.
    \item Integration of the uncertainty-aware depth completion network within a probabilistic volumetric
    mapping framework. 
    \item Evaluation of the framework on real-world outdoor datasets demonstrating deployment on a robot. %
\end{enumerate}

\section{Related Work} \label{S:related_work}
\subsection{Depth Completion}

Depth completion is the task of inferring a dense depth image from
a sparse one, with or without the guidance of a corresponding
camera image. Classical image processing techniques~\cite{ip-basic}
have been applied for unguided depth completion. While these approaches
require minimal computational resources, they do not exploit 
camera image texture, and typically under-perform image-guided methods.

Most recent works on guided depth completion are learning-based. Ma~\etal~\cite{sparse2dense} developed a depth completion network 
using ResNet~\cite{resnet} as a backbone, and later improved it by 
adopting a UNet~\cite{unet} encoder-decoder paradigm  
within a self-supervised training framework~\cite{Ma2019}. 
Additional information such as surface normals~\cite{deeplidar} and
semantics~\cite{completion_segmentation} have been shown to be useful for depth completion.
Optimizations based on learning affinity matrix~\cite{cspn, cspn++} have also
been proposed to further refine the predicted depth and improve accuracy.
In this work, we leverage the accuracy of such image-guided networks 
while keeping the network compact. 
This allows the system to be deployed on mobile GPUs.

Multimodal fusion of visual and depth sensing
using deep neural networks remains 
challenging.
Strategies include concatenating the two modalities to form a combined input (early fusion) as well as fusing features extracted from the two modalities at a later stage (late fusion) have been 
explored~\cite{completion_segmentation, deeplidar}. 
A further challenge arises when the networks are required to generalize to lidars with different
sparsity patterns and resolutions~\cite{sparseconv, normalisedconveldesokey}. 
In this paper, we keep the same sparsity level in input depth data during training and at test time.

\subsection{Depth Completion for 3D Reconstruction and Navigation}

While many works aim to address the problem of depth completion
on a per-image basis, only a few investigate the use of the dense depth output for applications such as reconstruction and navigation.
Fehr~\etal~\cite{completion_planning} proposed 
a planning system that predicts dense depth images using a CNN~\cite{sparse2dense}
and uses them as the input for the Voxblox mapping system~\cite{voxblox}. Depth 
estimation has been used in visual SLAM systems to recover metric 
scale~\cite{cnn_slam, drm_slam}, and to complete sparse visual 
features~\cite{zuo2021codevio, matsuki2021codemapping}.
In this work, we make depth completion more applicable for navigation by connecting it with a volumetric mapping system to achieve both denser reconstruction and to detect more free space.

An issue that often arises with depth completion networks trained on 
depth images generated by projecting 3D lidar points (e.g. KITTI~\cite{sparseconv}) is the invalid prediction for regions of the sky due to the lack of ground truth depth there.
As a consequence, directly using these predictions results in noisy 
reconstructions and incorrect identification of free space. In this work, we 
utilise synthetic datasets with sky annotations so that the network could learn 
to predict a very large depth value for the sky.

Uncertainty estimates, derived from heuristics~\cite{drm_slam} or learning~\cite{masha_completion}, have been used
to reject wrong predictions during reconstruction. 
Our work is closely related to~\cite{masha_completion}, 
which proposed a framework for depth completion and mapping
in indoor scenarios using RGB-D cameras. 
The authors designed a CNN to predict both dense depth and depth uncertainty, 
which are then
fed into an occupancy-based volumetric mapping system~\cite{supereight_funk}. 
The resulting occupancy maps uncover more 
\textit{obstacle-free} space compared to using the raw 
depth images and facilitate improved path planning. 
In our work, we extend these concepts
to real-world outdoor environments where the lidar depth data 
is much sparser than depth cameras and the scenes have a much larger scale.
In comparison to approaches that use RGB-D images as input, 
our method learns to complete missing information between the 
lidar points, rather than filling holes in depth images
which is characteristic of RGB-D images.

\section{Approach} \label{S:approach}

An overview of our proposed system for volumetric mapping with probabilistic depth completion
is shown in \figref{fig:pipeline}. The inputs to our system are
sets of grey-scale camera images and sparse lidar depth images. This input is processed by a
neural network with a shared encoder and two separate decoders to generate 
a completed dense depth image and the corresponding depth uncertainty 
prediction. We filter unreliable depth predictions considering their range 
and predicted uncertainty. The remaining predicted depth and 
uncertainty 
are then fed into 
an efficient probabilistic volumetric mapping system ~\cite{supereight_funk} 
to create a dense reconstruction.

\subsection{Network Architecture} \label{SS:network_architecture}
Our focus in
this paper is \textit{not} redesigning depth completion 
network architecture, as the performance on the KITTI Depth Completion 
Benchmark~\cite{sparseconv} is saturated, and accuracy comes with more 
computation burden. Rather, our aim is to obtain probabilistic depth 
completions applicable to robotic tasks including 3D reconstruction and path 
planning. Therefore, we choose three baseline depth completion networks from 
the KITTI Depth Completion Benchmark~\cite{sparseconv}: 1. \textit{\textbf{S2D}} 
\cite{sparse2dense}, a ResNet-based 
\cite{resnet} vanilla Encoder-Decoder network. 
 2. \textit{\textbf{U-S2D}}~\cite{Ma2019}, a ResNet-based Encoder-Decoder network with long skip-connections with concatenation.
3. \textit{\textbf{FastDepth}}*\footnote{We add one convolution layer for the depth image input, 
and the feature is concatenated with the first layer of the image feature as in 
\cite{Ma2019}.}~\cite{wofk2019fastdepth}, an efficient network with 
additive skip connection. Note that the network used in our framework is replaceable depending on the application.

In this work, we extend the depth completion network to incorporate 
depth uncertainty
prediction by adding an additional identical decoder to the chosen networks. 
For the uncertainty decoder in FastDepth*, we only keep the last three layers and one additive skip 
connection for efficiency.

\subsection{Loss Function and Training Strategy}
Our depth uncertainty network aims to predict aleatoric uncertainty
which comes from the noise inherent in the observations 
\cite{aleotoric_original, Kendall2017}. In the context of depth completion, 
depth at object boundaries usually has higher aleatoric uncertainty. 
Assuming the
depth measurement follows a
Gaussian distribution $\mathcal{N}(\mu,\sigma)$, the depth network is trained to estimate the mean $\mu$, while the uncertainty network estimates the standard 
deviation 
$\sigma$, the
aleatoric uncertainty.

The loss function~\cite{aleotoric_original} for jointly optimising the depth 
completion network and the depth
uncertainty network is defined as: 
\begin{equation}
L_{\text{unc}}=\frac{1}{N} \sum_{p=1}^{N} \frac{\left|y_{p}-f\left(\mathbf{x}_{p}\right)\right|^{2}}{\sigma\left(\mathbf{x}_{p}\right)^{2}}+\log \left(\sigma\left(\mathbf{x}_{p}\right)^{2}\right)
\label{loss_function}
\end{equation}
where $\mathbf{x}_{p}$ is the input vector of features for the depth 
completion network (sparse depth and camera image), $y_{p}$ is the 
ground truth depth measurement, $f(.)$ and $\sigma(.)$ are the 
completed depth and 
associated depth uncertainty output by the 
network.

The predicted uncertainty has an inverse component in the first term and a
proportional part in \eqref{loss_function}. This means that the 
uncertainty prediction should be optimised to be high at erroneous predictions, and low elsewhere. This uncertainty prediction 
is then used by 
the volumetric mapping system to update free space and reject incorrect depth predictions
which can improve the quality of both the reconstruction
and the free space estimation.

\subsection{Probabilistic Volumetric Mapping} \label{SS:mapping}

The predicted depth measurements are then integrated
into a volumetric reconstruction for use in robotic applications 
such as path planning. 
We use the state-of-the-art multi-resolution volumetric mapping pipeline~\cite{supereight_funk}
to leverage the explicitly represented free space. 
This pipeline demonstrates greater efficiency and lower memory usage
when integrating long-range lidar scans in large-scale 
environments~\cite{Wang2020}, compared to conventional 
occupancy mapping methods such as OctoMap~\cite{OctoMap13}.

For each voxel in the 3D reconstruction, 
we store its log-odds occupancy probability. 
Free, unknown and occupied space have
negative, zero and positive occupancy log-odds probabilities, 
respectively.
Computation of the log-odds occupancy probability $L(d)$
at distance $d$ along a ray $\mathbf{r}$ 
uses a piecewise linear function~\cite{supereight_funk}:
\begin{equation}
    L(d) =
    \begin{cases} 
        l_{\min} & d \leq 3 \sigma(d_r) \\
        \frac{-l_{\min}}{3 \sigma(d_r)} d & 3 \sigma(d_r) < d \leq \frac{k_{\tau} d_r}{2} \\
        \frac{-l_{\min}}{3 \sigma(d_r)} \frac{k_{\tau} d_r}{2} & \frac{k_{\tau} d_r}{2} < d \leq k_{\tau} d_r \\
        \text{no update} & \text{otherwise}
    \end{cases}
\end{equation}
where $l_{\min}$ denotes 
the minimum occupancy probability in log-odds, 
$d_r$ is the measured depth along a ray $\mathbf{r}$, 
$k_{\tau}$ is a scaling factor controlling 
how much occupied space is created behind a measured surface (surface 
thickness), and $\sigma(d_r)$ is the uncertainty 
of a measurement $d_r$. 

In the method in~\cite{masha_completion}, a quadratic model for 
$\sigma(d_r)$ is used,
which suited reconstruction with an RGB-D camera. In our work,
we estimate $\sigma(d_r)$ using a linear model
as:
\begin{equation}
	\sigma(d_r) =
	\begin{cases}
		\sigma_{\min}  & d_r\leq\frac{\sigma_{\min}}{k_{\sigma}} \\
		k_{\sigma} d_r & 
		\frac{\sigma_{\min}}{k_{\sigma}}<d_r<
		\frac{\sigma_{\max}}{k_{\sigma}}\\
		\sigma_{\max}  & \frac{\sigma_{\max}}{k_{\sigma}}\leq d_r\\
	\end{cases}
\label{linear_unc}
\end{equation}
to better represent the characteristics of a lidar.
$\sigma_{\min}$, $\sigma_{\max}$ and $k_{\sigma}$  
are constants that depend on the sensor specification.

Our mapping system utilises the depth-dependent uncertainty 
$\sigma(d_i)$ for 
raw depth $d_i$ from the input lidar measurement, and the 
network-predicted uncertainty for the predicted depth $d_p$, 
similar to~\cite{masha_completion}. Since our predicted uncertainty is trained to 
coincide with depth error,
we reject predicted depth when the predicted
uncertainty is higher than expected. Specifically, we invalidate 
a completed depth measurement if the predicted
uncertainty is more than 2 times greater than the sensor
uncertainty. Analysis of this rejection is presented in \secref{ablation_uncertainty}.
Unlike~\cite{masha_completion}, 
we did not update free
space for these uncertain depths. This is shown to remove more 
incorrect free space which is detrimental to safe navigation.

\section{Experimental Results} \label{S:experimental_results}

\subsection{Datasets}
\label{realworlddataset}
Depth completion is heavily studied for autonomous driving
~\cite{sparseconv}, but relatively under-explored in aerial, legged or UGV
platforms where the camera images have different depth distribution and objects 
than images captured on the street.
Thus, we evaluated our approach on two separate datasets that
target mobile robotics applications where explicit free space is required.

\subsubsection{Handheld Dataset}

The first dataset is a handheld-device dataset: the Newer College Dataset (NCD)~\cite{ramezani2020newer}. 
The dataset was collected using a 64-beam 3D lidar and 
stereoscopic-inertial camera at typical walking speeds. 
Specifically, we used the left grey-scale camera
images from an Intel Realsense D435i and lidar scans from 
a 64-beam Ouster lidar scanner, synchronised by VILENS~\cite{wisthunified,vilens}, 
as the input to our completion network. 
Sparse depth images were generated by projecting 3D lidar points onto the camera image plane. 
The dataset~\cite{ramezani2020newer} also provided a highly-accurate ground truth point cloud which is useful for evaluating the reconstruction. The ground truth trajectory of the handheld-device was provided by the dataset.
Poses of the handheld-device were computed by localizing individual scans against this ground truth map using VILENS.

Training the depth completion network requires ground truth depth.
We followed the approach of the KITTI Depth Completion Benchmark 
\cite{sparseconv} and accumulated 11 consecutive laser scans to increase
the density of the depth images and use this as the ground truth depth.
We simulate a sparser and lower-cost 16-beam lidar providing input to our 
network by downsampling the original 64-beam lidar 
readings by removing scan lines.

We generated 11087 samples from the long experiment in NCD and 6651 
samples from the short experiment. The long experiment samples are 
used for training, while 1000 samples from the short experiment are 
randomly selected for testing and the remaining 5651 samples for 
validation. 
The camera image, input and ground truth depth images all have the 
same resolution of 848 $\times$ 480 pixels.

\begin{figure}[t]
	\centering
	\includegraphics[width=0.8\columnwidth]{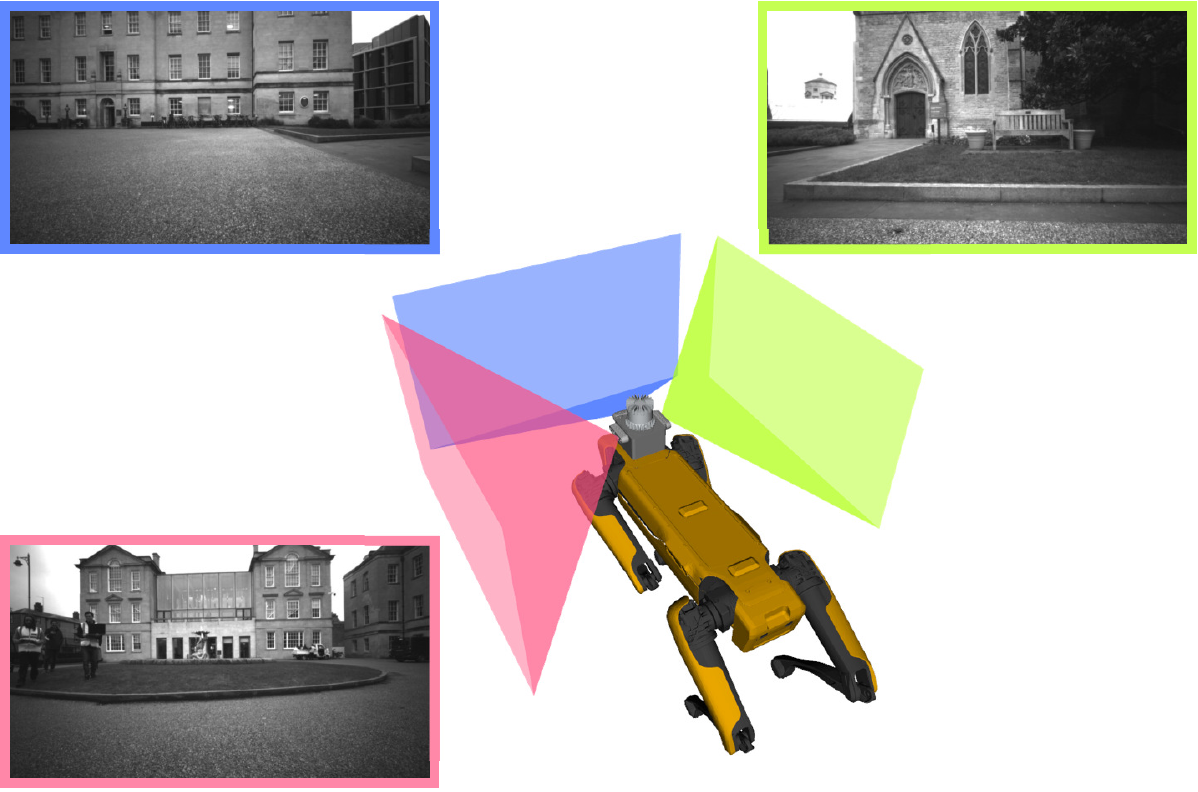}
	\caption{Our mapping device has cameras facing left, forward and right beneath a LIDAR and is mounted Spot robot. The multi-camera setup has an overlap of about 270\degree with the lidar.}
	\label{fig:spot}
\end{figure}

\subsubsection{Robot Dataset with three-camera setup}

Most lidar depth completion works use only a forward-facing monocular camera image. Thus, around three-fourths of the lidar points are not considered during depth completion. 
Instead, we built a device with three Intel Realsense D435i cameras and a 64-beam Ouster lidar scanner.
The three cameras were arranged as shown in \figref{fig:spot} facing front, left and right respectively.
This device was mounted on a legged robot, Boston Dynamics Spot.
Data was collected while the robot was manually operated around an outdoor campus environment.
Similar pre-processing as in the above Handheld Dataset was applied to generate the input and ground truth depth images, and the poses of the scanning device.
We refer to this dataset as the Mathematics Institute (Maths Inst.) dataset and use it solely for testing our proposed method.

\begin{figure*}[t]
	\centering
	\includegraphics[width=1.9\columnwidth]{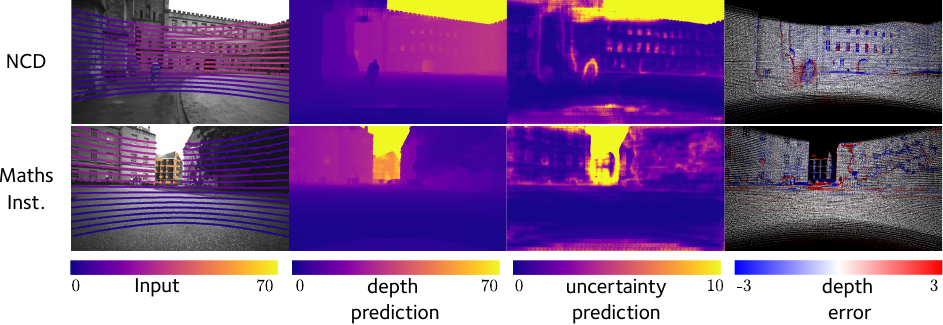}
	\caption{Example results from vKITTI rel + NCD unc, U-S2D in \tabref{network_evaluation}. Left to right: camera 
		image overlaid with sparse lidar depth, dense completed depth, uncertainty prediction, error in depth prediction. All measurements are in metres. Regions of high uncertainty correspond to regions of higher error in depth prediction, allowing us to reject these points for cleaner reconstructions.}
	\label{fig:predictions}
\end{figure*}

\subsubsection{Synthetic Pre-training Dataset}
In addition to NCD, we also use the Virtual KITTI 2 dataset (vKITTI)~\cite{vkitti2} for training the depth network.
This dataset contains photo-realistic image sequences captured 
from a simulated vehicle with ground truth depth images. The depth for regions of the sky is encoded as 256m, in contrast to real-world datasets like KITTI or NCD where the depth there is unannotated (not covered by lidar). Thus, the depth network can predict a pre-defined large depth in such regions. 

To generate the sparse input depth image, we applied a 16-beam lidar mask from 
NCD to the dense ground truth depth to keep input depth sparsity consistent 
with the above datasets, similar to~\cite{guidenet}. This is crucial for generalisation as discussed in \ref{generalisability}, since conventional convolution kernels are known to be sensitive to changes in the 
density of valid pixels in the input~\cite{sparseconv, completion_segmentation}.
We prepared a training set with 40520 images and a validation set with 1000 images.
Colour images were converted to grey-scale and cropped to have the same width as the images of the above datasets.

\subsection{Implementation Details} 
\subsubsection{Network Training Details}
Our implementation is built upon~\cite{aerial_completion}. The depth completion network follows the open-source implementation from~\cite{sparse2dense, Ma2019, wofk2019fastdepth}. 
We used the Adam optimiser with an initial learning rate of 
10e-4 and decrease it by a factor of 10 every 5 epochs. 
The training was performed on an NVIDIA Tesla V100 GPU.

Minimising the uncertainty loss function $L_{\text{unc}}$
in~\eqref{loss_function} corresponds to optimising the depth completion 
network and the depth uncertainty network simultaneously. 
We observed that this joint training does not always 
lead to optimal depth accuracy.
Hence, we experimented with a two-stage training procedure:
pre-training the shared encoder and the depth decoder with a selected loss function first,
and then training the uncertainty decoder using $L_{\text{unc}}$, while freezing the weights of the shared encoder and the depth decoder.
For each stage, we trained the network for 30 epochs and
selected the best-performing model based on the validation set. 
Even though this leads to sub-optimal performance on the uncertainty objective $L_{\text{unc}}$, the depth prediction is more important and is hence prioritised over uncertainty.

\subsubsection{Volumetric Mapping System}
We used the volumetric mapping system from Funk~\etal~\cite{supereight_funk} to fuse the 3D reconstruction using the lidar data and poses as described in \secref{SS:mapping}
with a voxel resolution of 6.5\,cm. The linear inverse sensor model is set such that surface thickness
 $k_{\tau}=0.026$, linear uncertainty model coefficient $k_{\sigma}=0.052$, $\sigma_{\min}=0.06$\,m, $\sigma_{\max}=0.20$\,m.

\subsection{Evaluation of the Depth Completion Network}\label{SS:dc_net}

We perform a quantitative evaluation of the network with
multiple training strategies and loss functions on two real-world datasets, as summarised in 
\tabref{network_evaluation}. Qualitative results are shown in \figref{fig:predictions}. We experimented with training on real-world (NCD) and/or synthetic datasets (vKITTI).

\begin{table*}[t]
    \caption{Depth prediction network performance under multiple training strategies,~\ref{SS:dc_net}.}
	\centering
	\begin{tabular}{l c c c c c c c c c c}
		\toprule

		\multirow{2}{0.06\linewidth}{Training Scheme}&\multirow{2}{0.06\linewidth}{Network} & \multirow{2}{0.06\linewidth}{\centering RMSE $\downarrow$ (m)} & \multirow{2}{0.06\linewidth}{\centering MAE $\downarrow$ (m)} & 
		\multirow{2}{0.06\linewidth}{\centering REL $\downarrow$ (\%)} &  \multirow{2}{0.06\linewidth}{\centering iMAE $\downarrow$ (km\textsuperscript{-1})}   &
		\multirow{2}{0.06\linewidth}{\centering $\delta_{1.05}$ $\uparrow$ (\%)}&
		\multirow{2}{0.06\linewidth}{\centering $\delta_{1.10}$ $\uparrow$ (\%)}&
		\multirow{2}{0.06\linewidth}{\centering $\delta_{1.25}$ $\uparrow$ (\%)}&
		 $L_{\text{unc}}$$\downarrow$ & AUSE $\downarrow$
		\\ \\
 		\hline
        \addlinespace
        \multicolumn{4}{l}{\textbf{NCD}}
	    
	    \\
	    \hline
		
		\addlinespace
		Linear Interpolation & -&3.96&1.51&16.4&-&68.4&77.2&85.8&-&-
		\\
		\hdashline
		\addlinespace
		NCD unc & S2D& \textbf{3.84} & 1.69 & 17.1 & 22.0 & 59.6 & 72.5 & 84.8& \textbf{0.88}&0.12
		\\

		NCD rel + NCD unc&S2D&4.58&1.76&10.8&24.5&52.4&76.0&88.1&1.52&0.16
		\\
		NCD rel + NCD unc & U-S2D&4.41&1.68&\textbf{\,9.9}&79.5&61.0&80.7&89.7&1.83&0.16
		\\
		\hdashline
		\addlinespace
		vKITTI rel + NCD unc & FastDepth* & 3.98&\textbf{1.37}&11.2&53.5&68.2&80.4&89.9&1.71&0.20
		\\
		vKITTI rel + NCD unc & S2D & 6.53&1.85&16.1&18.8&56.0&73.0&88.5&1.28&\textbf{0.11}
		\\
		vKITTI rel + NCD unc & U-S2D&5.21&1.42&11.5&\textbf{17.4}&\textbf{81.9}&\textbf{89.8}&\textbf{94.3}&1.29&0.14
		\\

		\midrule
        \textbf{Maths Inst.}
       	\\
        \hline
        \addlinespace\centering 
         Linear Interpolation &-&\textbf{2.69}&0.89&18.6&-&60.5&72.7&85.7&-&-
        \\
       	\hdashline
       	\addlinespace
         KITTI 16 rel & S2D & 3.21&1.28&27.7&84.4&37.0&53.4&73.6&-&-
        \\
        KITTI 64 rel  & S2D & 3.64&1.77&50.1&99.1&27.5&44.7&67.1&-&-
        \\
    	\hdashline
        \addlinespace
		
        NCD unc & S2D& 2.81&1.22&26.6&60.9&35.0&53.4&77.6&3.72&0.14
        
        \\
      	\hdashline
        \addlinespace
        
        vKITTI rel + NCD unc & FastDepth* & 2.79&\textbf{0.83}&\textbf{14.5}&53.0&58.8&73.7&88.4&\textbf{3.43}&0.22
        \\
        vKITTI rel + NCD unc & S2D & 4.22&1.08&19.6&45.7&50.2&67.2&86.4&8.8&\textbf{0.13}
        \\
        vKITTI rel + NCD unc & U-S2D&3.72&0.89&15.3&\textbf{41.5}&\textbf{62.3}&\textbf{75.9}&\textbf{88.8}&5.02&0.15
        
        \\
        \addlinespace

		\midrule
		\textbf{KITTI Validation Set}
		\\
		\hline
		\addlinespace
		KITTI 16-beam & S2D & 1.34&0.45&2.4&1.9&90.5&96.7&99.3&-&-

		\\
		KITTI 64-beam & S2D & \textbf{1.05}&\textbf{0.34}&\textbf{1.8}&\textbf{1.5}&\textbf{93.9}&\textbf{98.0}&\textbf{99.5}&-&-
		\\

		\bottomrule
        \addlinespace
        \end{tabular}
    
        \label{network_evaluation} 
\end{table*}

\subsubsection{Evaluation Metrics}

As an established dataset for depth completion evaluation, KITTI uses
RMSE as the primary metric for evaluation. 
However, RMSE and MAE penalise inaccurate depth predictions for distant objects heavily. 
In robotics applications, depth prediction for nearby objects is as important as those which are further away.
In order to balance accuracy across different depth ranges, we use mean absolute relative error (REL) as our primary loss function for the depth completion network, which is defined as 
\begin{equation}
	REL=\frac{1}{N} \sum_{p=1}^{N} 
	\left|\frac{y_{p}-f\left(\mathbf{x}_{p}\right)}{y_{p}}\right|
	\label{rel_error}.
\end{equation}

We also report the mean absolute error of the inverse depth (iMAE), and the percentage of predicted pixels whose relative error is less than a threshold $\delta \in \{1.05, 1.10, 1.25\}$.

For uncertainty metrics, in addition to our uncertainty loss function $L_{\text{unc}}$, we also evaluated uncertainty using the Area Under the 
Sparsification Error curve (AUSE)~\cite{Ilg2018}. This metric 
captures the correlation between the estimated uncertainty and 
prediction error. 

\subsubsection{Two-stage Training}
Training the depth and uncertainty networks together using the uncertainty loss 
$L_{\text{unc}}$ in NCD (NCD unc in \tabref{network_evaluation}) leads 
to good performance on $L_{\text{unc}}$ itself. However, there is no explicit 
objective for the depth network, and some metrics such as REL are not always 
optimal. We then experiment with a two-stage training strategy to separate the 
optimisation of depth and uncertainty estimation:
we train the depth network first with our chosen loss function
REL, and then train the uncertainty decoder using $L_{\text{unc}}$ (NCD rel + NCD unc).
Such two-stage training enables us to utilise both synthetic and 
real-world datasets: we
 train the depth network on vKITTI, and the 
 uncertainty decoder on NCD (vKITTI rel + NCD unc in \tabref{network_evaluation}), to get the best of both worlds: the depth network benefits from large training samples and precise depth (including the sky) in vKITTI and the uncertainty decoder captures real-world depth characteristics in NCD. We use U-S2D trained with this strategy for the reconstruction experiments, for its generally good performance on depth and uncertainty metrics.

\subsubsection{Generalisability}
\label{generalisability}
We train S2D on the KITTI Depth Completion dataset with raw 64-beam depth images 
and downsampled 16-beam depth images. However, the depth estimation is significantly worse when tested on the NCD/Maths Inst. datasets.
We hypothesise that the lack of generalisability is due to inconsistency in the 
input depth sparsity pattern between KITTI and our test datasets. 
 When we apply the same 
input depth mask from NCD 
to vKITTI as the training dataset, the depth network only trained on vKITTI  
generalises to both NCD and Maths Inst without any fine-tuning, validating our 
hypothesis and approach.  

\subsubsection{Runtime Analysis}

We test the three probabilistic depth completion networks on a mobile GPU 
Nvidia Quadro M2200, which resembles the resources of a typical mobile robot. 
The inference speed for FastDepth* is 9.7Hz, S2D 8.3Hz, and U-S2D 1.4 Hz, 
allowing for near real-time operation on a robot. Note that for a fair 
comparison, FastDepth* is not pruned and not hardware-optimised (TVM). This, along with the additional uncertainty decoder and larger image size, contributes to slower inference than
reported in~\cite{wofk2019fastdepth}.

\subsection{Evaluation of the Reconstruction and Free Space} 
We perform a quantitative evaluation of reconstruction quality and
the free space predicted by our proposed method (U-S2D, vKITTI rel + NCD unc 
from \tabref{network_evaluation}).
We compare reconstructions generated with the following configurations: 
\begin{enumerate}
    \item 16-beam lidar depth image, linear sensor uncertainty
    \item 64-beam lidar depth image, linear sensor uncertainty
    \item completed depth image, predicted uncertainty
\end{enumerate}

To evaluate the accuracy of free and occupied space
in the reconstruction, we create an occupancy map using
the centimetre-accurate 3D point cloud captured by a survey-grade
Leica laser scanner cloud~\cite{ramezani2020newer}. 
We generate the ground truth free and occupied space by performing ray casting from the sensor location to every point in the ground truth cloud.
The ground truth map uses the same resolution as our generated reconstructions. 

\begin{table}[!h]
	\caption{\small{Evaluation using different depth images for mapping.}}
	\centering
	\begin{tabular}{ l l l r r l}
		\toprule
		&   &\multicolumn{2}{c}{Reconstruction} & 	\multicolumn{2}{c}{Free Space}
		\\ 
		Section & Type &  \multirow{2}{0.6cm}{Error (m)}& \multirow{2}{0.6cm}{Vol. (m\textsuperscript{3})} & 
        \multirow{2}{1cm}{Correct  (m\textsuperscript{3})} & \multirow{2}{0.8cm}{Incorrect} \\ \\
		\hline
		\addlinespace
		\multirow{3}{1cm}{NCD}&  raw 16 (5 Hz)&0.08 &25.5 &6025.7 &1.55\%  
		\\
		& raw 64 & 0.09& 52.0& 12916.1 & 1.50\%
		\\
		& completed mono& 0.17& 100.7 & 10479.2 & 0.85\%
		\\
		\hline
		\addlinespace
		\multirow{3}{1cm}{Maths Inst.}
		&  raw 16 & 0.12&28.4&3675.1&2.3\%
		\\
		
		& raw 64 &0.13&46.7&6798.6&2.9\%
		\\

		& completed mono&0.15&23.9&2791.8&2.2\%
		\\
		& completed 3-cam&0.14&37.8&5416.5&2.8\%
		\\
		\bottomrule
        \addlinespace
	\end{tabular}
    \footnotesize{Rejection ratio $\rho=2$ is used for completed depth based reconstruction in NCD. We used lidar scan at 5 Hz. Only depth within 50m is integrated.}
	\label{free_space}
\end{table}

The quantitative results are summarised in \tabref{free_space}.
 We analyse the following two aspects:

\subsubsection{Reconstruction}

\begin{figure*}[]
    \centering
    \includegraphics[width=2\columnwidth]{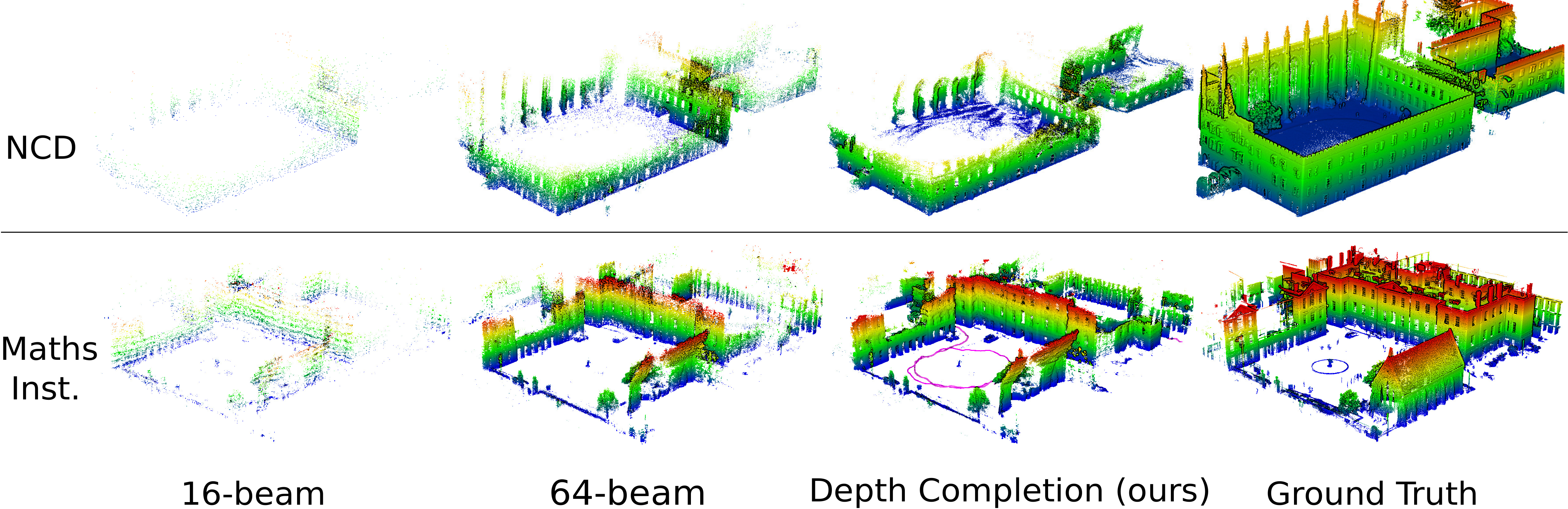}
    \caption{Mesh reconstruction using different inputs for NCD with a monocular camera (top row) and Maths Inst. dataset with multi-cameras (bottom row). The trajectory followed by the robot in the Maths Inst. dataset is shown in the third image on the bottom row.}
    \label{fig:mappings}
    \vspace{-4mm}
\end{figure*}

The reconstruction is evaluated in terms of accuracy and 
completeness. For accuracy, we create meshes using marching cubes 
(\figref{fig:mappings})
from the volumetric map and sample points from them. Then, we 
compute the average distance from the sampled points to the ground truth point cloud. 
For completeness, we calculate the volume of occupied space 
in the ground truth occupancy map that overlaps with our reconstruction. 
Our reconstruction using completed depth recovers significantly more
ground truth occupied space than the reconstruction with raw 16 beams,
while retaining an average point-to-point error below 0.2\,m.

\subsubsection{Free Space}
We compare the free space detected by our reconstruction to free space reconstructions derived from occupancy maps produced from the ground truth point clouds of both NCD and Maths Inst. datasets.
A free voxel is identified as being incorrect free space if it is free in the reconstruction but either occupied or unknown in the ground truth. 
We observe that the reconstruction using our approach with completed depth reveals much more correct free space than when using raw 16-beam lidar depth. The performance is similar to using the full dense 64-beam lidar. We also demonstrate the advantage of depth completion using the three-camera setup compared to only upsampling with a single camera.
The three-camera configuration recovers around twice the amount ($\sim$5416$\text{m}^{\text{3}}$) of free space as compared to the single camera configuration ($\sim$2791$\text{m}^{\text{3}}$)
The incorrect free space from our method is within 2.5\% of the total discovered free space, which is important for safe navigation.

\subsection{Effect of Uncertainty Rejection}

\label{ablation_uncertainty}
\begin{figure}[]
	\centering
	\includegraphics[width=\columnwidth]{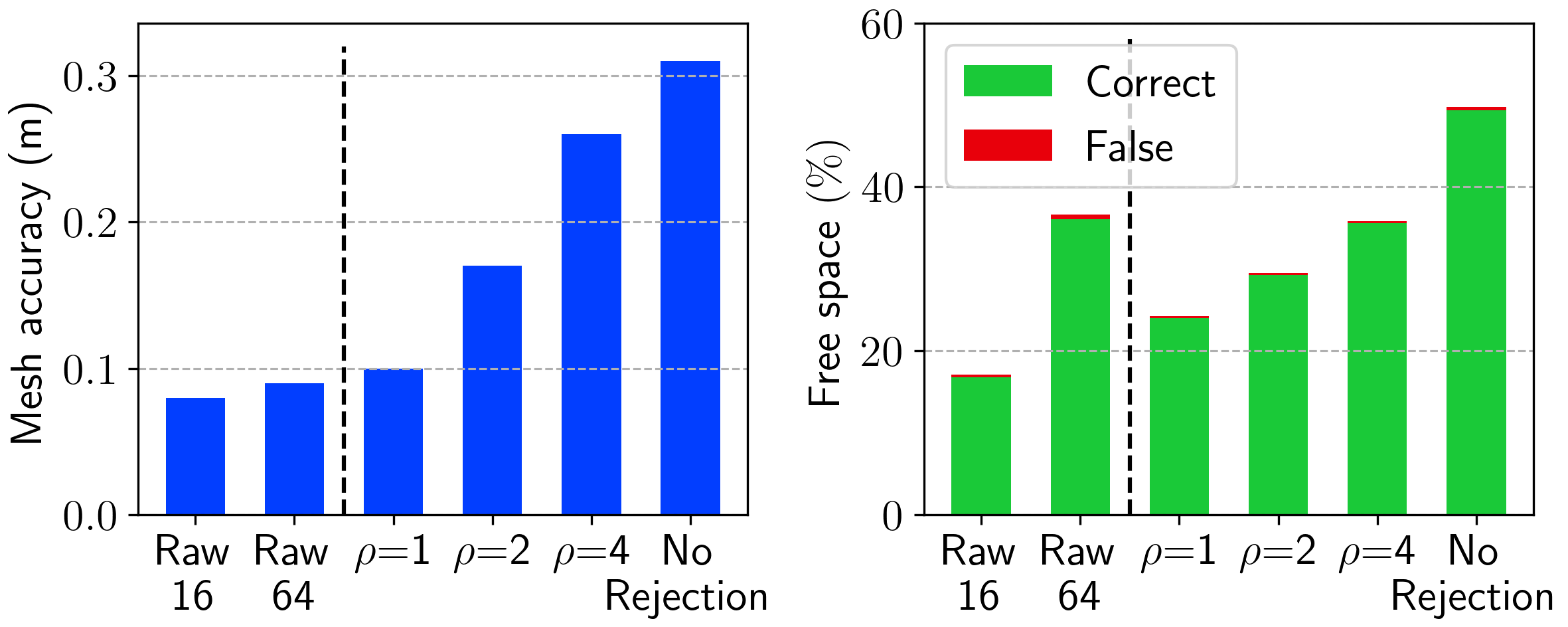}
	\caption{Performance for different uncertainty rejection ratios.}
	\label{fig:uncertainty_plot}
\end{figure}

\begin{figure}[]
	\centering
	\includegraphics[width=0.9\columnwidth]{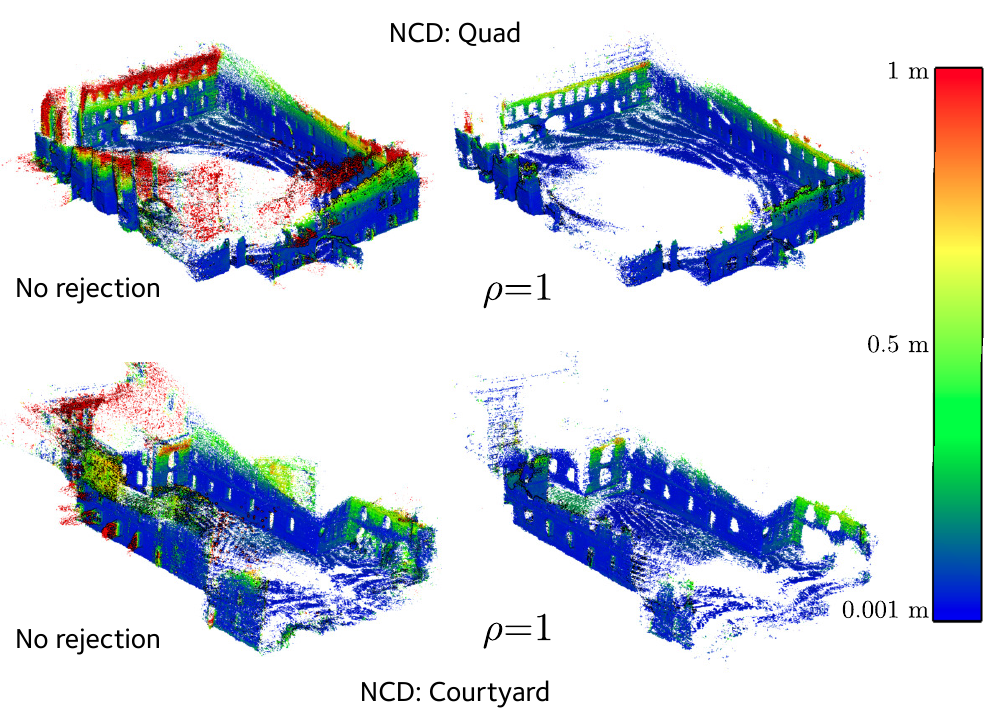}
	\caption{Example mesh reconstructions using NCD with varying uncertainty rejection thresholds. Point-to-point errors 
		are visualised with a colour map. Without any uncertainty rejection, we have a noisier reconstruction coloured in red.}
	\label{fig:uncertainty_mapping}
	\vspace{-4mm}
\end{figure}

The uncertainty prediction from our network estimates the standard 
deviation of the predicted depth. Here, we present an ablation 
study on the effect of uncertainty rejection. We only update free space and 
surface information when the predicted uncertainty is less than $\rho$ times
the expected linear uncertainty given by~\eqref{linear_unc}.
For NCD, we tested different 
choices of $\rho$ and calculate the mesh accuracy, percentage of correct
free space and wrong free space. We used the sparse 16-beam lidar depth as
a mapping baseline. As shown in \figref{fig:uncertainty_plot},
both the reconstruction error and correct space increase with $\rho$, as expected.
The detected free space is comparable to the raw 64 beam sensor for $\rho=4$.
In our experiments, we used $\rho=2$ for a balance between reconstruction
accuracy and free space detection.
We illustrate the effect of uncertainty rejection in \figref{fig:uncertainty_mapping}. 

\subsection{Path Planning}

\begin{figure}[t]
    \centering
    \includegraphics[width=\columnwidth]{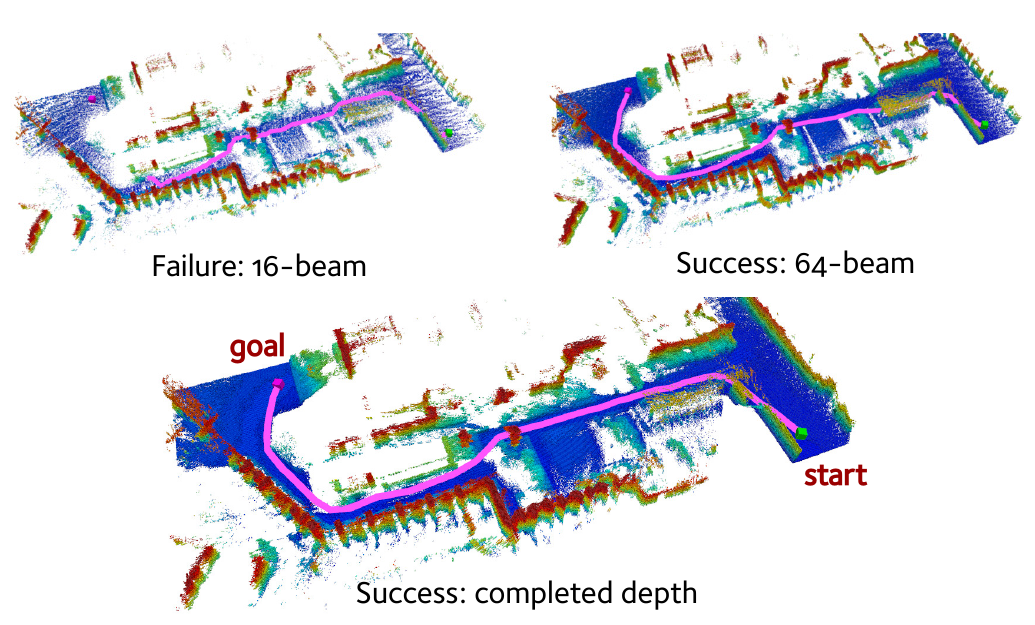}
    \caption{Planning using a map built with 16 beam data is sparse and fails to find a path, whereas
             our approach using probabilistic completed depth makes it feasible.}
    \label{fig:planning}
\end{figure}

Our final experiments demonstrate the advantage of using an occupancy map, with probabilistic depth completion, for path planning.
We illustrate planning results using 64-beam, 16-beam lidar and completed depth as inputs respectively. We use the RRT* planner~\cite{Karaman2011} to generate a path
in the map that passes through narrow garden passages between two buildings. 
As demonstrated in \figref{fig:planning}, RRT* fails to find a path around a tight corner
in the map created using 16-beam lidar, as its sparse measurement cannot detect sufficient free space. 
In contrast, the much denser reconstruction from our approach allows RRT* to find a path in the same scenario.

\section{Conclusions and Future Work} \label{S:conclusions}

In summary, we propose a 3D lidar reconstruction framework with 
probabilistic depth completion, which achieves denser reconstructions
as well as more free space detection compared to using sparse lidar depth images.
Such an approach enables the use of low-cost sparse lidar scanners to achieve performances similar to expensive scanners with higher beam density. 
We demonstrate that our approach is suitable for deployment on real-world robots for applications such as 3D reconstruction and path planning.

The proposed approach facilitates map compression while preserving dense reconstructions. We would like to analyse this capability for tasks such as localization and navigation in the future.
Extending the depth completion approach with wide field-of-view cameras, such as fisheye cameras is also of interest.

\section*{Acknowledgment}
The authors would like to thank David Wisth, Milad Ramezani and Matias Mattamala for their help on generating the training datasets for depth completion, Michal Staniaszek for his assistance with operating the Spot robot.

\bibliographystyle{IEEEtran}
\footnotesize
\bibliography{2022-iros-tao}

\begin{thebibliography}{10}
\providecommand{\url}[1]{#1}
\csname url@rmstyle\endcsname
\providecommand{\newblock}{\relax}
\providecommand{\bibinfo}[2]{#2}
\providecommand\BIBentrySTDinterwordspacing{\spaceskip=0pt\relax}
\providecommand\BIBentryALTinterwordstretchfactor{4}
\providecommand\BIBentryALTinterwordspacing{\spaceskip=\fontdimen2\font plus
\BIBentryALTinterwordstretchfactor\fontdimen3\font minus
  \fontdimen4\font\relax}
\providecommand\BIBforeignlanguage[2]{{%
\expandafter\ifx\csname l@#1\endcsname\relax
\typeout{** WARNING: IEEEtran.bst: No hyphenation pattern has been}%
\typeout{** loaded for the language `#1'. Using the pattern for}%
\typeout{** the default language instead.}%
\else
\language=\csname l@#1\endcsname
\fi
#2}}

\bibitem{sparseconv}
J.~Uhrig, N.~Schneider, L.~Schneider, U.~Franke, T.~Brox, and A.~Geiger,
  ``Sparsity invariant {CNNs},'' \emph{Intl. Conf. on 3D Vision (3DV)}, p.
  11–20, 2017.

\bibitem{sparse2dense}
F.~Ma and S.~Karaman, ``Sparse-to-dense: Depth prediction from sparse depth
  samples and a single image,'' in \emph{IEEE Intl. Conf. on Robotics and
  Automation (ICRA)}, 2018, pp. 4796--4803.

\bibitem{Ma2019}
F.~Ma, G.~V. Cavalheiro, and S.~Karaman, ``Self-supervised sparse-to-dense:
  Self-supervised depth completion from {LiDAR} and monocular camera,'' in
  \emph{IEEE Intl. Conf. on Robotics and Automation (ICRA)}, 2019, pp.
  3288--3295.

\bibitem{wofk2019fastdepth}
D.~Wofk, F.~Ma, T.-J. Yang, S.~Karaman, and V.~Sze, ``Fastdepth: Fast monocular
  depth estimation on embedded systems,'' in \emph{IEEE Intl. Conf. on Robotics
  and Automation (ICRA)}.\hskip 1em plus 0.5em minus 0.4em\relax IEEE, 2019,
  pp. 6101--6108.

\bibitem{cspn++}
X.~Cheng, P.~Wang, C.~Guan, and R.~Yang, ``{CSPN++}: Learning context and
  resource aware convolutional spatial propagation networks for depth
  completion,'' \emph{AAAI Conf. on Artificial Intelligence}, vol.~34, no.~07,
  pp. 10\,615--10\,622, Apr. 2020.

\bibitem{penet}
M.~Hu, S.~Wang, B.~Li, S.~Ning, L.~Fan, and X.~Gong, ``{PENet}: Towards precise
  and efficient image guided depth completion,'' in \emph{IEEE Intl. Conf. on
  Robotics and Automation (ICRA)}, 2021.

\bibitem{ip-basic}
J.~Ku, A.~Harakeh, and S.~L. Waslander, ``In defense of classical image
  processing: Fast depth completion on the {CPU},'' \emph{Conf. on Computer and
  Robot Vision (CRV)}, p. 16–22, 2018.

\bibitem{resnet}
K.~He, X.~Zhang, S.~Ren, and J.~Sun, ``Deep residual learning for image
  recognition,'' in \emph{Proc. {IEEE} Int. Conf. Computer Vision and Pattern
  Recognition}, 2016, pp. 770--778.

\bibitem{unet}
O.~Ronneberger, P.~Fischer, and T.~Brox, ``{U-Net}: Convolutional networks for
  biomedical image segmentation,'' in \emph{Medical Image Computing and
  Computer-Assisted Intervention (MICCAI)}, 2015, pp. 234--241.

\bibitem{deeplidar}
J.~Qiu, Z.~Cui, Y.~Zhang, X.~Zhang, S.~Liu, B.~Zeng, and M.~Pollefeys,
  ``{DeepLiDAR}: Deep surface normal guided depth prediction for outdoor scene
  from sparse {LiDAR} data and single color image,'' in \emph{Proc. {IEEE} Int.
  Conf. Computer Vision and Pattern Recognition}, 2019, pp. 3313--3322.

\bibitem{completion_segmentation}
M.~Jaritz, R.~d. Charette, E.~Wirbel, X.~Perrotton, and F.~Nashashibi, ``Sparse
  and dense data with {CNNs}: Depth completion and semantic segmentation,''
  \emph{Intl. Conf. on 3D Vision (3DV)}, p. 52–60, 2018.

\bibitem{cspn}
X.~Cheng, P.~Wang, and R.~Yang, ``Depth estimation via affinity learned with
  convolutional spatial propagation network,'' in \emph{Eur. Conf. on Computer
  Vision (ECCV)}, 2018, pp. 103--119.

\bibitem{normalisedconveldesokey}
A.~Eldesokey, M.~Felsberg, and F.~S. Khan, ``Confidence propagation through
  {CNNs} for guided sparse depth regression,'' \emph{{IEEE} Trans. Pattern
  Anal. Machine Intell.}, vol.~42, no.~10, pp. 2423--2436, 2020.

\bibitem{completion_planning}
M.~{Fehr}, T.~{Taubner}, Y.~{Liu}, R.~{Siegwart}, and C.~{Cadena},
  ``{Predicting Unobserved Space for Planning via Depth Map Augmentation},'' in
  \emph{IEEE Intl. Conf. on Robotics and Automation (ICRA)}, 2019, pp. 30--36.

\bibitem{voxblox}
H.~Oleynikova, Z.~Taylor, M.~Fehr, R.~Siegwart, and J.~Nieto, ``Voxblox:
  Incremental {3D} euclidean signed distance fields for on-board {MAV}
  planning,'' \emph{IEEE/RSJ Intl. Conf. on Intelligent Robots and Systems
  (IROS)}, p. 1366–1373, 2017.

\bibitem{cnn_slam}
K.~Tateno, F.~Tombari, I.~Laina, and N.~Navab, ``{CNN-SLAM}: Real-time dense
  monocular {SLAM} with learned depth prediction,'' \emph{Proc. {IEEE} Int.
  Conf. Computer Vision and Pattern Recognition}, p. 6565–6574, 2017.

\bibitem{drm_slam}
X.~Ye, X.~Ji, B.~Sun, S.~Chen, Z.~Wang, and H.~Li, ``{DRM-SLAM}: Towards dense
  reconstruction of monocular {SLAM} with scene depth fusion,''
  \emph{Neurocomputing}, vol. 396, p. 76–91, 2020.

\bibitem{zuo2021codevio}
X.~Zuo, N.~Merrill, W.~Li, Y.~Liu, M.~Pollefeys, and G.~Huang, ``Codevio:
  Visual-inertial odometry with learned optimizable dense depth,'' in
  \emph{IEEE Intl. Conf. on Robotics and Automation (ICRA)}.\hskip 1em plus
  0.5em minus 0.4em\relax IEEE, 2021, pp. 14\,382--14\,388.

\bibitem{matsuki2021codemapping}
H.~Matsuki, R.~Scona, J.~Czarnowski, and A.~J. Davison, ``Codemapping:
  Real-time dense mapping for sparse slam using compact scene
  representations,'' \emph{{IEEE} Robotics and Automation Letters}, vol.~6,
  no.~4, pp. 7105--7112, 2021.

\bibitem{masha_completion}
M.~Popovi\'{c}, F.~Thomas, S.~Papatheodorou, N.~Funk, T.~Vidal-Calleja, and
  S.~Leutenegger, ``Volumetric occupancy mapping with probabilistic depth
  completion for robotic navigation,'' \emph{{IEEE} Robotics and Automation
  Letters}, vol.~6, no.~3, pp. 5072--5079, 2021.

\bibitem{supereight_funk}
N.~Funk, J.~Tarrio, S.~Papatheodorou, M.~Popovi\'{c}, P.~F. Alcantarilla, and
  S.~Leutenegger, ``Multi-resolution {3D} mapping with explicit free space
  representation for fast and accurate mobile robot motion planning,''
  \emph{{IEEE} Robotics and Automation Letters}, vol.~6, no.~2, pp. 3553--3560,
  2021.

\bibitem{aleotoric_original}
D.~Nix and A.~Weigend, ``Estimating the mean and variance of the target
  probability distribution,'' in \emph{Proceedings of 1994 IEEE International
  Conference on Neural Networks (ICNN'94)}, vol.~1, 1994, pp. 55--60 vol.1.

\bibitem{Kendall2017}
A.~Kendall and Y.~Gal, ``{What uncertainties do we need in Bayesian deep
  learning for computer vision?}'' in \emph{Conf. on Neural Information
  Processing Systems (NIPS)}, 2017, pp. 5575--5585.

\bibitem{Wang2020}
Y.~Wang, N.~Funk, M.~Ramezani, S.~Papatheodorou, M.~Popovi\'{c}, M.~Camurri,
  S.~Leutenegger, and M.~Fallon, ``{Elastic and Efficient LiDAR Reconstruction
  for Large-Scale Exploration Tasks},'' in \emph{IEEE Intl. Conf. on Robotics
  and Automation (ICRA)}, 2021.

\bibitem{OctoMap13}
A.~Hornung, K.~M. Wurm, M.~Bennewitz, C.~Stachniss, and W.~Burgard,
  ``{OctoMap}: An efficient probabilistic {3D} mapping framework based on
  octrees,'' \emph{Autonomous Robots}, vol.~34, no.~3, pp. 189--206, Apr. 2013.

\bibitem{ramezani2020newer}
M.~{Ramezani}, Y.~{Wang}, M.~{Camurri}, D.~{Wisth}, M.~{Mattamala}, and
  M.~{Fallon}, ``The {Newer College Dataset}: Handheld {LiDAR}, inertial and
  vision with ground truth,'' in \emph{IEEE/RSJ Intl. Conf. on Intelligent
  Robots and Systems (IROS)}, 2020.

\bibitem{wisthunified}
D.~Wisth, M.~Camurri, S.~Das, and M.~Fallon, ``Unified multi-modal landmark
  tracking for tightly coupled lidar-visual-inertial odometry,'' \emph{{IEEE}
  Robotics and Automation Letters}, vol.~6, no.~2, pp. 1004--1011, 2021.

\bibitem{vilens}
D.~Wisth, M.~Camurri, and M.~Fallon, ``{VILENS}: Visual, inertial, lidar, and
  leg odometry for all-terrain legged robots,'' \emph{arXiv preprint
  arXiv:2107.07243}, 2021.

\bibitem{vkitti2}
Y.~Cabon, N.~Murray, and M.~Humenberger, ``Virtual kitti 2,'' 2020.

\bibitem{guidenet}
J.~Tang, F.-P. Tian, W.~Feng, J.~Li, and P.~Tan, ``Learning guided
  convolutional network for depth completion,'' \emph{{IEEE} Trans. on Image
  Processing}, vol.~30, p. 1116–1129, 2021.

\bibitem{aerial_completion}
L.~Teixeira, M.~R. Oswald, M.~Pollefeys, and M.~Chli, ``Aerial single-view
  depth completion with image-guided uncertainty estimation,'' \emph{{IEEE}
  Robotics and Automation Letters}, vol.~5, no.~2, p. 1055–1062, 2019.

\bibitem{Ilg2018}
E.~Ilg, O.~Cicek, S.~Galesso, A.~Klein, O.~Makansi, F.~Hutter, and T.~Brox,
  ``Uncertainty estimates and multi-hypotheses networks for optical flow,'' in
  \emph{Eur. Conf. on Computer Vision (ECCV)}, 2018, pp. 652--667.

\bibitem{Karaman2011}
S.~Karaman and E.~Frazzoli, ``Sampling-based algorithms for optimal motion
  planning,'' \emph{Intl. J. of Robotics Research}, vol.~30, no.~7, pp.
  846--894, 2011.

\end{thebibliography}

\end{document}